\documentclass[10pt, a4paper]{article}
\usepackage{amsmath}
\usepackage{graphicx}
\usepackage{subcaption}
\usepackage{amssymb}
\usepackage{bm} 
\usepackage[final]{lrec2026} 

\title{Graph Fusion Across Languages using Large Language Models}

\name{Kaung Myat Kyaw$^{*}$\thanks{$^{*}$ Equal contribution.}, Khush Agarwal$^{*}$, Jonathan Chan} 

\address{Innovative Cognitive Computing Research Center (IC2) \\
School of Information Technology, KMUTT \\ 
         \{kaungmyat.kyaw, khush.agar\}@kmutt.ac.th,  jonathan@sit.kmutt.ac.th\\
         }
         
\abstract{
Combining multiple knowledge graphs (KGs) across linguistic boundaries is a persistent challenge due to semantic heterogeneity and the complexity of graph environments. We propose a framework for cross-lingual graph fusion, leveraging the in-context reasoning and multilingual semantic priors of Large Language Models (LLMs). The framework implements structural linearization by mapping triplets directly into natural language sequences (e.g., [head] [relation] [tail]), enabling the LLM to map relations and reconcile entities between an evolving fused graph ($G_{c}^{(t-1)}$) and a new candidate graph ($G_{t}$). Evaluated on the DBP15K dataset, this exploratory study demonstrates that LLMs can serve as a universal semantic bridge to resolve cross-lingual discrepancies. Results show the successful sequential agglomeration of multiple heterogeneous graphs, offering a scalable, modular solution for continuous knowledge synthesis in multi-source, multilingual environments.  Our implementation and experimental framework are publicly available in our anonymized repository: \url{https://anonymous.4open.science/r/KG-Fusion-1A7D/README.md}
 \\ \newline  }

\begin{document}

\maketitleabstract

\section{Introduction}

The landscape of Artificial Intelligence has been fundamentally reshaped by the emergence of Large Language Models (LLMs)~\cite{grattafiori2024llama, yang2025qwen3, team2023gemini}. These models, pre-trained on expansive, multilingual corpora, have surpassed their initial role as statistical text predictors to become sophisticated reasoning engines. By internalizing the semantic relationships across hundreds of languages, LLMs have demonstrated an unprecedented capacity for cross-lingual zero-shot transfer \cite{chirkova2024zero}. In parallel, Knowledge Graphs (KGs) have remained the foundational bedrock of structured, symbolic AI \cite{hogan2021knowledge}. Unlike the probabilistic nature of neural models, KGs provide a deterministic and interpretable representation of real-world facts, serving as the ground truth for domain-specific applications such as biomedical research, legal compliance, and global financial analysis \cite{ji2021survey, peng2023knowledge}.

As the global digital ecosystem becomes increasingly interconnected, a critical challenge has emerged in the management of multilingual knowledge graphs \cite{perevalov2024multilingual}. Knowledge is naturally fragmented across linguistic and cultural boundaries \cite{huang2022multilingual}; for example, the English-centric part of DBpedia \cite{lehmann2015dbpedia} may offer extensive coverage of Western historical events, while its Chinese or Japanese counterparts provide far greater granularity regarding Eastern heritage and regional entities. The task of cross-lingual entity alignment (EA) and graph fusion is therefore essential to bridge this information, facilitating the creation of a truly universal knowledge base. Historically, this task has been dominated by embedding-based strategies \cite{yang2019aligning} or complex Graph Neural Networks (GNNs) \cite{zhang2023cross, tong2022joint}. However, these traditional methods are often constrained by their reliance on substantial seed data, which comprises manually pre-aligned entity pairs used to learn mapping functions between disparate vector spaces \cite{huo2024zeroea}. Such requirements create a significant barrier to entry, particularly for low-resource languages or emerging domains where seed alignments are non-existent.

\begin{figure}[t]
\begin{center}
\includegraphics[scale=0.365]{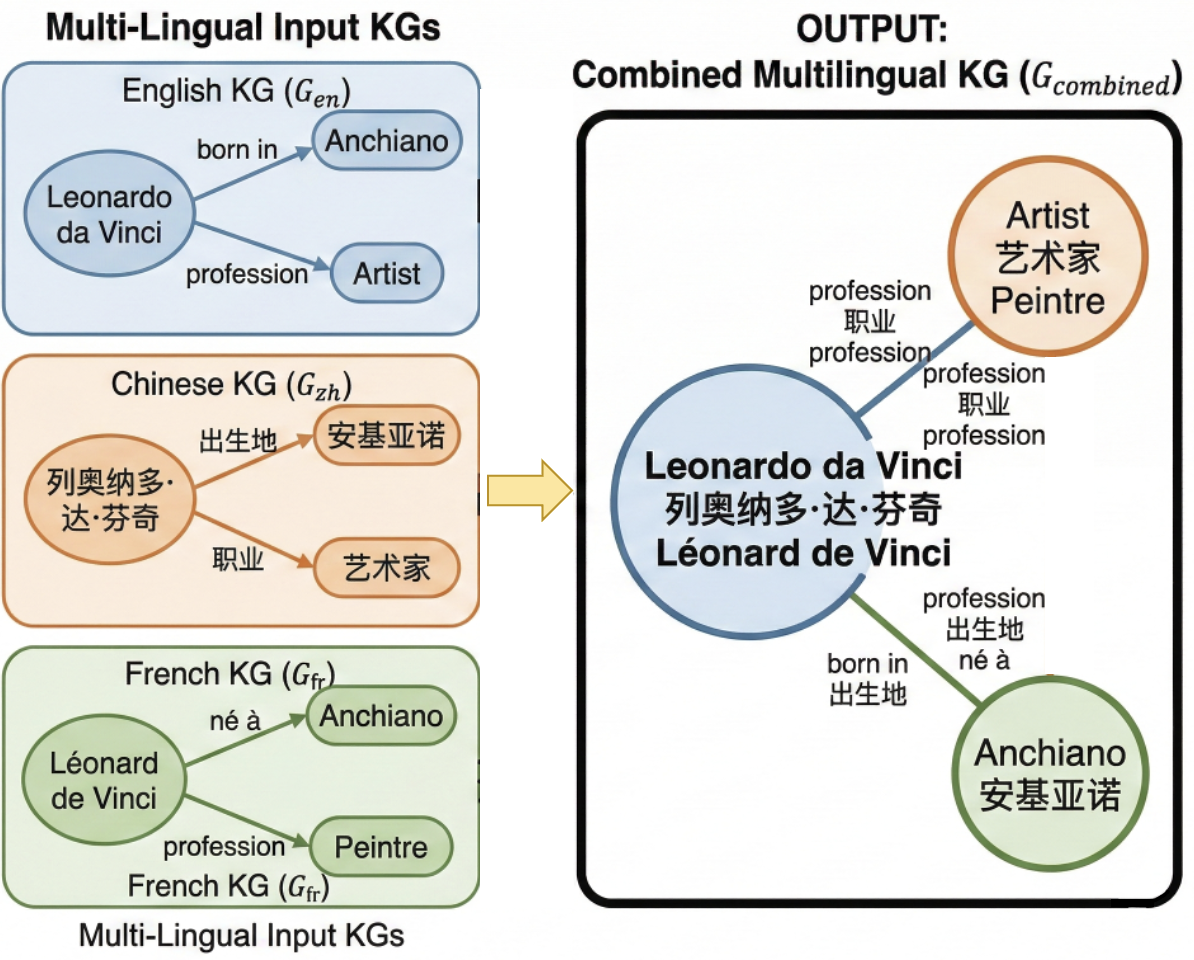} 
\caption{Conceptual framework for LLM-based cross-lingual knowledge graph fusion. Heterogeneous input graphs in English ($G_{en}$), Chinese ($G_{zh}$), and French ($G_{fr}$) are processed through a fusion framework that linearizes structural triplets into natural language and the LLM acts as a semantic bridge to reconcile entities.}
\label{fig.1}
\end{center}
\end{figure}

Despite the promise of leveraging LLMs for this task, the field currently faces a critical structural-semantic alignment bottleneck when attempting to scale to $N$-graph scenarios \cite{chen2016multilingual}. The primary challenge is the proliferation of \textbf{cross-lingual discrepancies}—a multifaceted issue encompassing lexical gaps (e.g., disparate writing systems and character sets), schema drift (where independent graphs utilize divergent relational ontologies), and variations in expressive granularity across languages. As we transition from binary alignment to the sequential agglomeration of $N$ graphs, these discrepancies become exponentially complex. For example, a relation such as \textit{``capital\_of''} in English might lack a direct 1-to-1 lexical equivalent in a third or fourth language, which may instead utilize a broader or more culturally specific relational label. Traditional string-matching and rigid embedding techniques \cite{liu2022selfkg} are inherently brittle when faced with these nuanced semantic overlaps. Furthermore, the scalability-context trade-off poses a fundamental architectural challenge: KGs are non-linear, high-dimensional structures, whereas LLMs operate on linear sequences within finite context windows.

\begin{figure*}[!ht]
\begin{center}
\includegraphics[width=\textwidth]{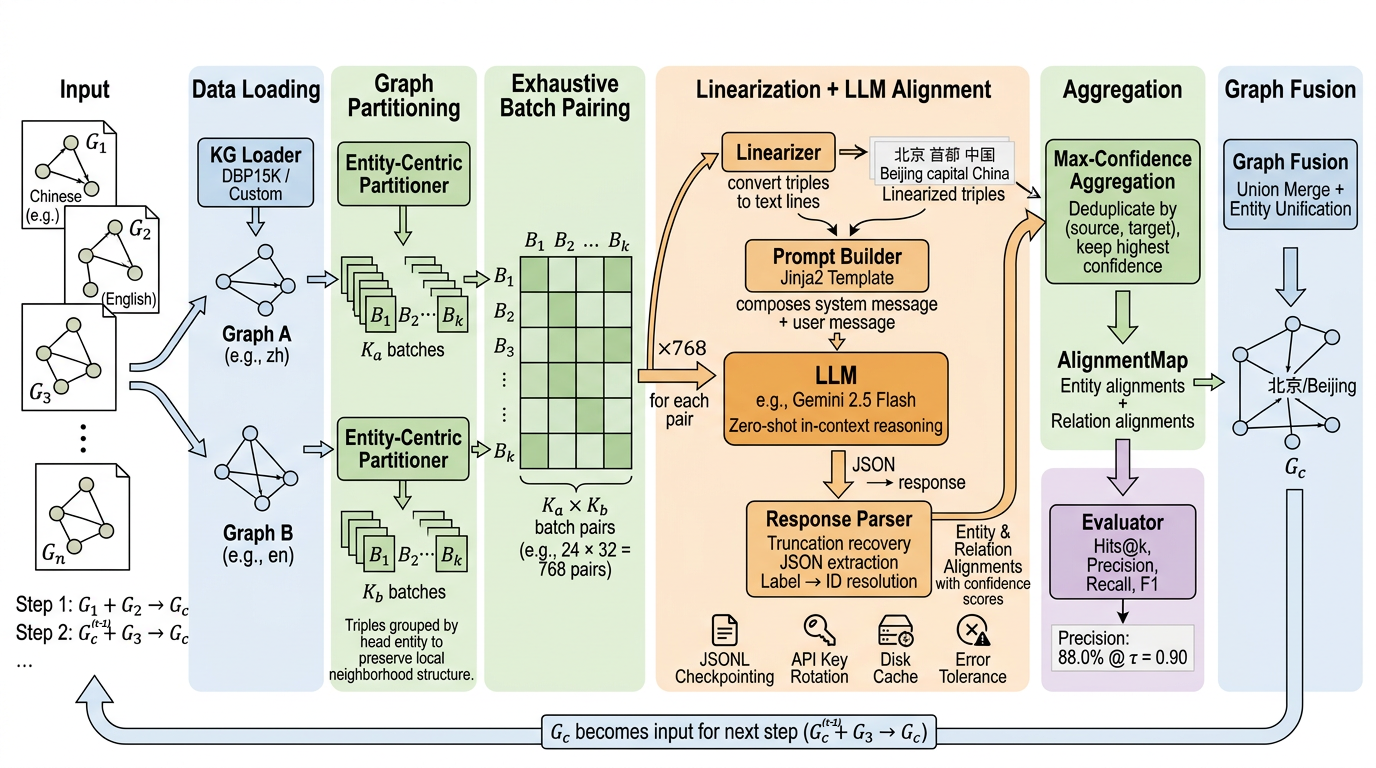} 
\caption{Overview of the proposed modular framework for $N$-Graph Knowledge Graph Fusion.}
\label{fig.2}
\end{center}
\end{figure*}

The process of flattening a graph into text which is a necessity for LLM consumption risks the loss of critical ``topological neighborhood'' context. Without observing the surrounding nodes and edges, an LLM may struggle to disambiguate entities with identical names but different structural roles \cite{jin2024large, zhu2024llms, pan2024unifying}. Most current alignment systems \cite{zhao2020experimental} are developed as specialized ``black boxes'' optimized for specific model architectures or fixed language pairs, lacking a modular infrastructure that allows researchers to swap LLM or test various partitioning strategies in a dynamic, plug-and-play fashion.

In this paper, we propose a novel, modular framework for LLM-based $N$-Graph Knowledge Graph Fusion. As illustrated in Figure~\ref{fig.1}, we move away from static mappings and position the LLM as a flexible semantic glue that discovers entity and relation alignments through sequential agglomeration and in-context reasoning. An overview of this modular architecture is illustrated in Figure~\ref{fig.2}. Our work makes several key contributions to the field of knowledge integration. 
\begin{itemize}
    \item We introduce a sequential agglomeration fusion pipeline based on a ``rolling'' strategy, where $N$ graphs are sequentially integrated into an evolving global graph, $G_{combined}$. This prevents the computational explosion typically associated with multi-graph matching. 
    \item To address scalability, we implement a context-aware graph partitioner that subdivides large KGs into manageable, readable triplet batches, ensuring the LLM maintains a local structural view of the entities it is aligning. 
\end{itemize}
We provide an evaluation on the multilingual DBP15K \cite{sun2017cross}, covering Chinese-English, Japanese-English, and French-English pairs. Our results demonstrate that by optimizing prompt structures and linearization formats, our pipeline achieves superior precision in cross-lingual mapping, proving that LLMs can serve as the primary engine for large-scale, multilingual graph synthesis.

\section{Preliminaries \& Related Works}
A KG is defined as a directed multigraph $G = (E, R, T)$, where $E$ is the set of entities, $R$ is the set of relation types, and $T \subseteq E \times R \times E$ is a set of triplets $(h, r, t)$. In a Multilingual KG (M-KG) setting, we consider $N$ disparate graphs $\{G_1, G_2, \dots, G_n\}$ where each $G_i$ represents knowledge in a specific language $L_i$. The objective of Cross-Lingual Entity Alignment (EA) is to find a set of alignment pairs $A_{i,j} = \{(e_i, e_j) \in E_i \times E_j \mid e_i \equiv e_j\}$, where $\equiv$ denotes semantic equivalence of real-world entities across linguistic barriers. Historically, this process relies on seed alignments (or "seeds") which is a small set of manually pre-aligned entity pairs that serve as the ground-truth supervision \cite{sun2017cross, zhao2020experimental}. These seeds act as anchors to bridge disparate graphs; however, they are often expensive to curate, creating a bottleneck for low-resource languages where such gold standard links are non-existent.

This task was addressed through knowledge representation learning and translational models such as TransE \cite{bordes2013translating}, which assumes that for any valid triplet, the embeddings $\mathbf{h, r, t} \in \mathbb{R}^d$ satisfy $\mathbf{h} + \mathbf{r} \approx \mathbf{t}$. To bridge linguistic gaps, models like MTransE \cite{chen2016multilingual} and JAPE \cite{sun2017cross} introduced transition matrices $\mathbf{M}_{ij}$ to map an entity $\mathbf{e}_i$ from $G_i$ into the vector space of $G_j$, such that $\|\mathbf{M}_{ij}\mathbf{e}_i - \mathbf{e}_j\| \to 0$. However, these distance-based approaches are highly sensitive to the quality and volume of seed alignments, as pre-aligned entity pairs used as supervision are often scarce or non-existent in low-resource or emerging domains. While GNNs such as GCN-Align \cite{wang2018cross} and RDGCN \cite{wu2019relation} utilize neighborhood information to disambiguate entities, they are language-blind without high-quality initial word embeddings. Moreover, GNNs are computationally expensive for large-scale graphs and often operate as "black boxes," making it difficult to interpret why an alignment was made or to adapt to new languages without retraining.

With the rise of models like Llama 3 \cite{grattafiori2024llama} and Gemini \cite{team2023gemini}, the field has shifted from distance-based matching to in-context reasoning. Recent frameworks like ZeroEA \cite{huo2024zeroea} and Seg-Align \cite{yang2024advancing} demonstrate that LLMs can perform alignment with zero or minimal seed data by treating entities as text. Models like ProLEA \cite{munne2025entity} generate "contextual profiles" for entities, transforming structured triplets into natural language summaries. This "semantic layer" bridges the gap between different linguistic representations.

While binary alignment ($G_1 \to G_2$) is well-studied, $N$-Graph fusion remains a frontier. MultiEA \cite{yang2024aligning} attempts to align multiple KGs in a single pass by clustering equivalent entities around a collective mean in a shared feature space. However, in true multilingual $N$-graph scenarios, the fusion process faces transitive error propagation: if $G_1$ is aligned to $G_2$ and $G_2$ to $G_3$, any semantic drift in the first pair is magnified in the second. Furthermore, current literature lacks a modular approach that can handle the semantic heterogeneity of $N$ languages simultaneously without incurring a computational cost.

By combining the structured reliability of KGs with the flexible, multi-step reasoning abilities of large language models, this work outlines a practical path toward a scalable, multilingual knowledge base. Rather than treating knowledge as fixed binary mappings, our approach supports an evolving synthesis of information that can grow across languages and domains. In this way, the study aims to help move the field toward more adaptive and globally integrated knowledge systems.

\section{Methodology}

\begin{figure*}[ht]
    \centering
    \begin{subfigure}{0.45\textwidth}
        \centering
        \includegraphics[width=\linewidth]{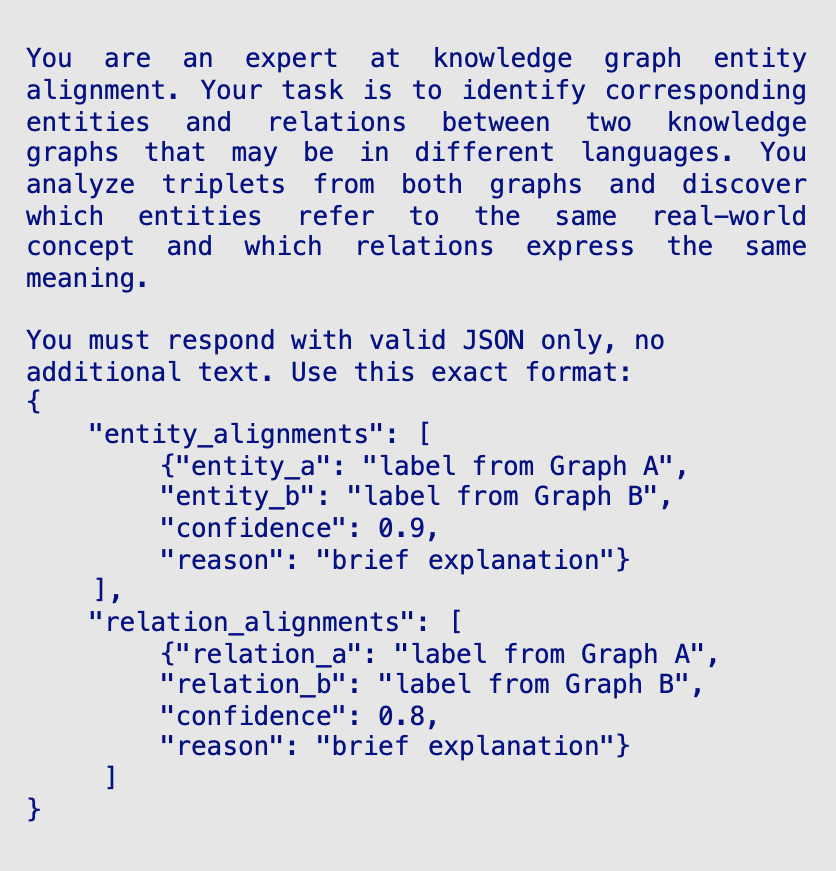}
        \caption{System Prompt}
    \end{subfigure}
    \hfill
    \begin{subfigure}{0.465\textwidth}
        \centering
        \includegraphics[width=\linewidth]{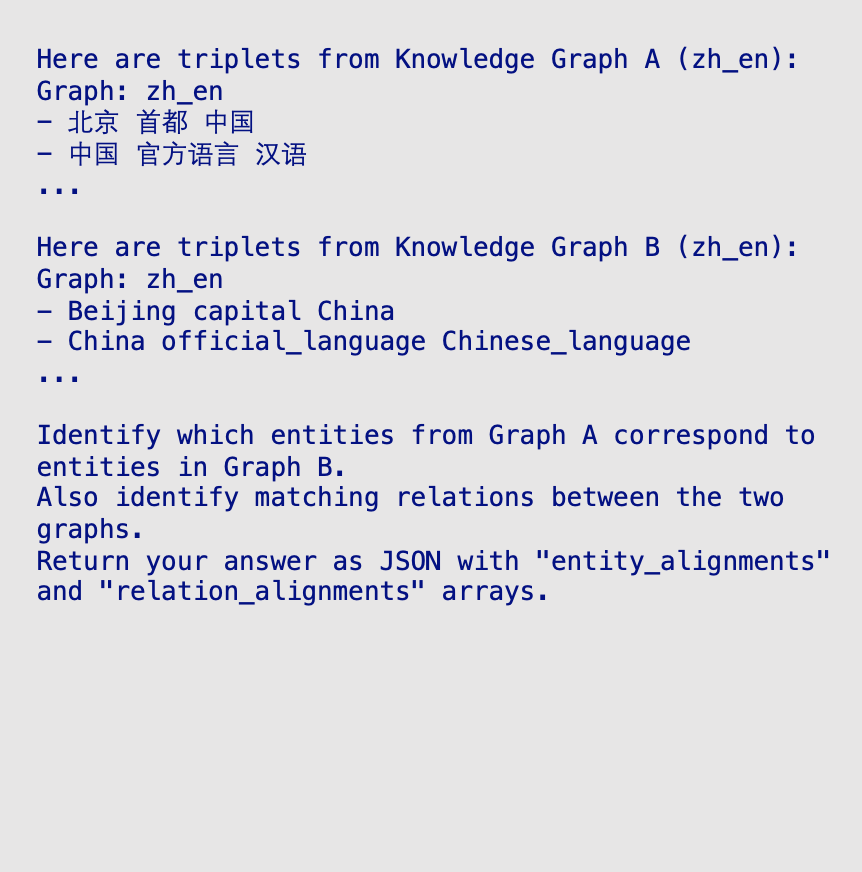}
        \caption{User Prompt}
    \end{subfigure}

    \caption{Details of the LLM prompting interface. The system prompt (left) defines the operational constraints and confidence thresholds, while the user prompt (right) serves as the data payload, containing the linearized triples from the source and target knowledge graphs.}
\label{fig:prompt}
\end{figure*}

\subsection{Problem Formulation}
We define the Multilingual $N$-Graph Fusion task as the sequential agglomeration of a global, unified knowledge graph $G_{c}^{(N)}$ from a set of $N$ heterogeneous graphs $\{G_1, G_2, \dots, G_n\}$ expressed in different languages. At each iteration $t \in [2, N]$, the framework seeks to align a new candidate graph $G_t$ with the previously synthesized global state $G_{c}^{(t-1)}$. The objective is to find an optimal alignment set $$A_t = \{(u_i, u_j) \mid u_i \in S_{c}^{(t-1)}, u_j \in S_t\}$$ where $S \in \{E, R\}$ that maximizes the joint probability of entity correspondences across linguistic boundaries. We formalize the decision for any entity or relation pairs $(u_i, u_j)$ via the objective function:
$$\max \sum_{(u_i, u_j) \in A_t} P(u_i \equiv u_j \mid \mathcal{L}(G_{c}^{(t-1)}), \mathcal{L}(G_t)),$$
where $\equiv$ denotes semantic equivalence, $\mathcal{L}$ is the linearization function. An alignment is accepted into the unified graph if its confidence score $\sigma$ satisfies:$$\sigma(u_i, u_j) \geq \tau,$$where $\tau$ is a predefined threshold (e.g., $\tau = 0.90$ for high-precision synthesis).

\subsection{Entity-Centric Graph Partitioning}
To manage the high dimensionality of $G$ and the finite context window $W$ of the LLM, we implement an entity-centric partitioning strategy. Given a set of triplets $T = \{(h, r, t)\}$, we partition $T$ into $k$ batches $\{B_1, B_2, \dots, B_k\}$ such that each batch $B_m$ is formed by grouping triples by their head entity $h$:$$B_m = \{ (h, r, t) \in T \mid h \in E_{sub} \},$$where $E_{sub}$ is a subset of entities assigned to the $m$-th partition. This ensures that the local topological neighborhood of an entity is preserved within a single processing unit, which is critical for structural disambiguation.

\subsection{Exhaustive Batch Pairing}
To ensure that every potential entity correspondence is evaluated, we implement an Exhaustive Batch Pairing strategy. Given the partitioned batch sets $\{B_1, \dots, B_k\}$ from $G_{c}^{(t-1)}$ and $\{B'_1, \dots, B'_{k'}\}$ from $G_t$, we construct a Cartesian product of all possible batch pairs:$$\mathcal{P} = \{ (B_i, B'_j) \mid 1 \leq i \leq k, 1 \leq j \leq k' \}$$This results in $k \times k'$ discrete reasoning tasks. While computationally intensive, this exhaustive approach guarantees that every entity neighborhood in the source graph is compared against every neighborhood in the target graph, maximizing the discovery of cross-lingual semantic anchors without relying on pre-computed candidate selection or heuristic blocking.

\subsection{Linearization and LLM Reasoning}
The core alignment engine operates via a Linearization Function $\mathcal{L}(B)$, which transforms structural triplets into a plain-text sequence $\mathcal{S}$. For each triplet $(h, r, t) \in B$, the transformation is defined as:$$\mathcal{L}(h, r, t) \to \text{"[head\_label] [relation\_label] [tail\_label]"}$$The resulting strings from a specific batch pair $(B_i, B'_j)$ are concatenated into a prompt $P$ consisting of a system persona $\pi$ and the linearized data. The prompts used for the LLM can be seen in Figure~\ref{fig:prompt}. The LLM performs zero-shot reasoning to discover a local mapping set $\hat{A}_{i,j}$ by maximizing the posterior probability of the alignment sequence:$$\hat{A}_{i,j} = \arg\max_{\hat{A}} P(\hat{A} \mid \mathcal{L}(B_i), \mathcal{L}(B'_j), \pi)$$where $B_i \subseteq G_{c}^{(t-1)}$ and $B'_j \subseteq G_t$. This generative approach allows the model to leverage its internal cross-lingual semantic priors to bridge the gap between disparate surface forms without requiring explicit seed translations.

\subsection{Robust Response Parsing and Resolution}
Due to the probabilistic nature of Large Language Model (LLM) generation, we implement a multi-stage Response Parser to handle failure modes such as output truncation or syntax errors. The parsing function $f_{parse}$ recovers partial JSON objects by identifying the last complete element in the sequence and closing unclosed braces to salvage alignment items that would otherwise be lost to token limits. The resolved textual labels are subsequently mapped back to internal dataset identifiers via a case-insensitive lookup function $f_{res}: \text{Label} \to \text{ID}$. To ensure global consistency, we apply a Max-Confidence Aggregation rule to deduplicate predictions appearing across multiple batch pairs. For any source entity or relation $u_i$, the final alignment $u_j^*$ is determined by:$$u_j^* = \text{arg max}_{u_j} \{ \sigma(u_i, u_j) \mid (u_i, u_j, \sigma) \in \hat{A}_{total} \}$$This aggregation effectively filters low-confidence noise and ensures that only the most probable semantic correspondences which of those that appear with the highest model certainty across various structural contexts are integrated into the unified global graph.

\subsection{Agglomeration Fusion and Schema Unification}
The framework performs a dual-track unification of the entity and relation sets. For every identified pair in $A_t$, the framework collapses the language-specific nodes into a single unified node that inherits a multilingual label set and all incident edges. To maintain schema consistency, we perform Relation Folding, where aligned predicates are merged into a consistent cross-lingual representation. The updated graph $G_{c}^{(t)}$ then serves as the base for the next iteration, maintaining a linear computational complexity $O(N)$ relative to the number of graphs.

\section{Experiments and Results}
This section provides a detailed quantitative and qualitative analysis of our LLM-based $N$-Graph fusion framework. We evaluate the system's ability to discover cross-lingual correspondences in a zero-shot environment, focusing on the trade-offs between precision, recall, and model confidence.

\subsection{Experimental Setup}
To assess the framework's efficacy, we utilize the DBP15K (Chinese-English) \cite{sun2017cross} dataset, which serves as the gold standard for cross-lingual entity alignment.
\begin{itemize}
    \item Dataset Composition: The $zh\_en$ pair consists of two knowledge graphs with 15,000 ground-truth alignment pairs.
    \item Zero-Shot Evaluation: Unlike traditional supervised methods, our approach requires no training data. We evaluate against the full 15,000 pairs (combining both training and test sets) to measure the system's total discovery capability.
    \item Model Configuration: The pipeline employs Gemini 2.5 Flash with the temperature set to 0.0 to ensure reasoning consistency and minimize hallucinatory variations.
    \item Runtime and Infrastructure: The pipeline completes the exhaustive batch processing in approximately 5–6 hours.
\end{itemize}

\subsection{Quantitative Performance Metrics}
The system demonstrated a robust ability to identify semantic equivalencies without any prior exposure to the dataset. Table~\ref{tab:primary_metrics} outlines the primary results for the $zh\_en$ language pair.

\begin{table}[ht]
\centering
\caption{Primary Alignment Metrics (DBP15K $zh\_en$)}
\label{tab:primary_metrics}
\begin{tabular}{lr}
\hline
\textbf{Metric} & \textbf{Value} \\ \hline
Total Batch Pairs & 768 \\
Unique Alignment Predictions & 5,416 \\
True Positives (TP) & 3,543 \\
False Positives (FP) & 1,873 \\
\textbf{Precision} & \textbf{65.4\%} \\
\textbf{Recall} & \textbf{23.6\%} \\
\textbf{F1 Score} & \textbf{34.7\%} \\ \hline
\end{tabular}
\end{table}

The model's recall is currently bounded by its total prediction volume; the system generated 5,416 unique alignment correspondences out of the 15,000 possible ground-truth pairs. Despite this coverage bottleneck, the precision of the generated predictions remains robust at 65.4\%, demonstrating strong initial performance for a purely zero-shot configuration.

\subsection{Hits@k Analysis and Source Accuracy}
The Hits@k metrics provide insight into the model's ranking capabilities, with Hits@1 measuring whether the correct target entity is identified as the top prediction. 

\begin{table}[ht]
\centering
\caption{Hits@k Evaluation}
\label{tab:hits_k}
\begin{tabular}{lrrr}
\hline
\textbf{k} & \textbf{Hits / 15,000} & \textbf{Rate} & \textbf{Acc. Predicted} \\ \hline
1 & 3,516 & 23.4\% & \textbf{88.3\%} \\
5 & 3,543 & 23.6\% & 89.0\% \\
10 & 3,543 & 23.6\% & 89.0\% \\ \hline
\end{tabular}
\end{table}

As detailed in Table~\ref{tab:hits_k}, when the model generates a prediction for a source entity, it identifies the correct target with an accuracy of \textbf{88.3\%} at Hits@1. The minimal performance gap between Hits@1 and Hits@5 (a 0.7 percentage point difference) further underscores the model's high precision in its top-ranked predictions.

\subsection{Impact of Confidence Filtering}
The model’s internal confidence score ($\sigma$) serves as a strong quality discriminator. True Positives exhibited a mean confidence of \textbf{0.980}, whereas False Positives averaged \textbf{0.738}.

\begin{table}[ht]
\centering
\setlength{\tabcolsep}{4pt} 
\caption{Confidence Threshold ($\tau$) Sensitivity Sweep}
\label{tab:threshold_sweep}
\begin{tabular}{lrrrrr}
\hline
\textbf{$\tau$} & \textbf{Pred.} & \textbf{TP} & \textbf{Prec.} & \textbf{Rec.} & \textbf{F1} \\ \hline
0.00 & 5,416 & 3,543 & 65.4\% & 23.6\% & 34.7\% \\
0.80 & 4,309 & 3,540 & 82.2\% & 23.6\% & 36.6\% \\
\textbf{0.90} & \textbf{4,002} & \textbf{3,520} & \textbf{88.0\%} & \textbf{23.5\%} & \textbf{37.0\%} \\
0.95 & 3,483 & 3,219 & 92.4\% & 21.4\% & 34.8\% \\ \hline
\end{tabular}
\end{table}

As illustrated in Table~\ref{tab:threshold_sweep}, by applying a threshold of $\tau = 0.90$, the framework achieves a precision of 88.0\% with negligible recall loss compared to the baseline.

\section{Conclusion}
This study introduced a modular, zero-shot framework for multilingual $N$-graph fusion leveraging the in-context reasoning capabilities of Large Language Models. By implementing an entity-centric partitioning strategy and a max-confidence aggregation rule, we successfully mitigated the challenges of structural heterogeneity and transitive error propagation that typically plague $N$-graph scenarios. Our results on the DBP15K dataset demonstrate that while recall remains a challenge due to context window limitations, the precision of LLM-based alignments, particularly when filtered by a 0.90 confidence threshold, is high. This suggests that LLMs can indeed serve as a universal semantic bridge for synthesizing global knowledge bases without the need for expensive, manually curated seed alignments.

\section{Future Work}
The current framework establishes a robust baseline for zero-shot $N$-graph fusion, yet several avenues for optimization remain to transition from exhaustive processing to more informed, selective integration. 

A primary direction involves the transition from exhaustive batch pairing to a multi-stage heuristic blocking and semantic indexing strategy. By utilizing lightweight multilingual embedding models to pre-index entity neighborhoods, the framework could perform candidate selection, directing LLM reasoning only toward high-probability pairs. This refinement would reduce the computational overhead of the Cartesian product while simultaneously improving recall by focusing resources on semantically ambiguous clusters.

Furthermore, the integration of global structural context within the linearization process presents an opportunity to move beyond localized triplet-level views. Future iterations could employ hierarchical prompting to provide the LLM with a schema-level overview or a summary of high-degree hub nodes prior to batch processing. Such an approach would likely mitigate disambiguation errors for entities with identical labels but distinct ontological roles. The stability of the sequential agglomeration may also be enhanced by investigating curated integration sequences, where graphs are merged based on linguistic proximity or graph density to minimize transitive semantic drift.

Beyond architectural shifts, exploring hybrid alignment architectures offers a path toward self-improving systems. LLM-discovered correspondences could serve as "silver seeds" to bootstrap and fine-tune traditional Graph Neural Network (GNN) aligners, combining the zero-shot reasoning of generative models with the structural efficiency of symbolic representations. Finally, evaluating the framework across a broader spectrum of evolving models will be essential to assess how variations in multilingual pre-training and context window management influence the precision of large-scale, cross-lingual knowledge synthesis.

\section{Bibliographical References}\label{sec:reference}

\bibliographystyle{lrec2026-natbib}
\bibliography{lrec2026-example}
\bibliographystylelanguageresource{lrec2026-natbib}
\bibliographylanguageresource{languageresource}

\end{document}